\documentclass[11pt,a4paper]{article}
\usepackage[hyperref]{emnlp2020}
\usepackage{times}
\usepackage{latexsym}

\usepackage{color}
\usepackage{graphicx}
\usepackage{amsfonts}
\usepackage{enumitem,kantlipsum,amsmath}

\usepackage{url}
\usepackage{microtype}
\usepackage{graphicx, multirow}
\newcommand{\ignore}[1]{}

\aclfinalcopy 

\title{Improving AMR Parsing with Sequence-to-Sequence Pre-training}

\author{Dongqin Xu$^1$\hspace{0.8cm}Junhui Li$^1$\thanks{Corresponding Author: Junhui Li.}\hspace{0.8cm}Muhua Zhu$^2$\hspace{0.8cm}Min Zhang$^1$\hspace{0.8cm} Guodong Zhou$^1$\\
$^1$School of Computer Science and Technology, Soochow University, Suzhou, China\\
$^2$Tencent News, Beijing, China\\
{xdqck@live.com},
{\{lijunhui, minzhang, gdzhou\}@suda.edu.cn}\\
{muhuazhu@tencent.com}
}

\date{}

\begin{document}
\maketitle
\begin{abstract}
In the literature, the research on abstract meaning representation (AMR) parsing is much restricted by the size of human-curated dataset which is critical to build an AMR parser with good performance. To alleviate such data size restriction, pre-trained models have been drawing more and more attention in AMR parsing. However, previous pre-trained models, like BERT, are implemented for general purpose which may not work as expected for the specific task of AMR parsing. In this paper, we focus on sequence-to-sequence (seq2seq) AMR parsing and propose a seq2seq pre-training approach to build pre-trained models in both single and joint way on three relevant tasks, i.e., machine translation, syntactic parsing, and AMR parsing itself. Moreover, we extend the vanilla fine-tuning method to a multi-task learning fine-tuning method that optimizes for the performance of AMR parsing while endeavors to preserve the response of pre-trained models. Extensive experimental results on two English benchmark datasets show that both the single and joint pre-trained models significantly improve the performance (e.g., from 71.5 to 80.2 on AMR 2.0), which reaches the state of the art. The result is very encouraging since we achieve this with seq2seq models rather than complex models. We make our code and model available at \url{https://github.com/xdqkid/S2S-AMR-Parser}.
\end{abstract}

\section{Introduction}

Abstract meaning representation (AMR) parsing aims to translate a textual sentence into a directed and acyclic graph which consists of concept nodes and edges representing the semantic relations between the nodes~\cite{banarescu_etal_aclws_2013}. Previous studies focus on building diverse approaches to modeling the structure in AMR graphs, such as tree-based approaches~\cite{wang_etal_naacl_2015,groschwitz_etal_acl_2018}, graph-based approaches~\cite{flanigan_etal_acl_2014,werling_etal_acl_2015,cai_lam_emnlp_2019}, transition-based approaches~\cite{damonte_etal_eacl_2017,guo_lu_emnlp_2018}, sequence-to-sequence (seq2seq) approaches~\cite{peng_etal_eacl_2017,noord_bos_2017,konstas_etal_acl_2017,ge_etal_ijcai_2019}, and sequence-to-graph (seq2graph) approaches~\cite{zhang_etal_acl_2019,zhang_etal_emnlp_2019,cai_lam_acl_2020}. Among these approaches, seq2seq-based approaches, which properly transform AMR graphs into sequences, have received much interest, due to the simplicity in implementation and the competitive performance.

\begin{figure}[t]
\begin{center}
\includegraphics[width=3.0in]{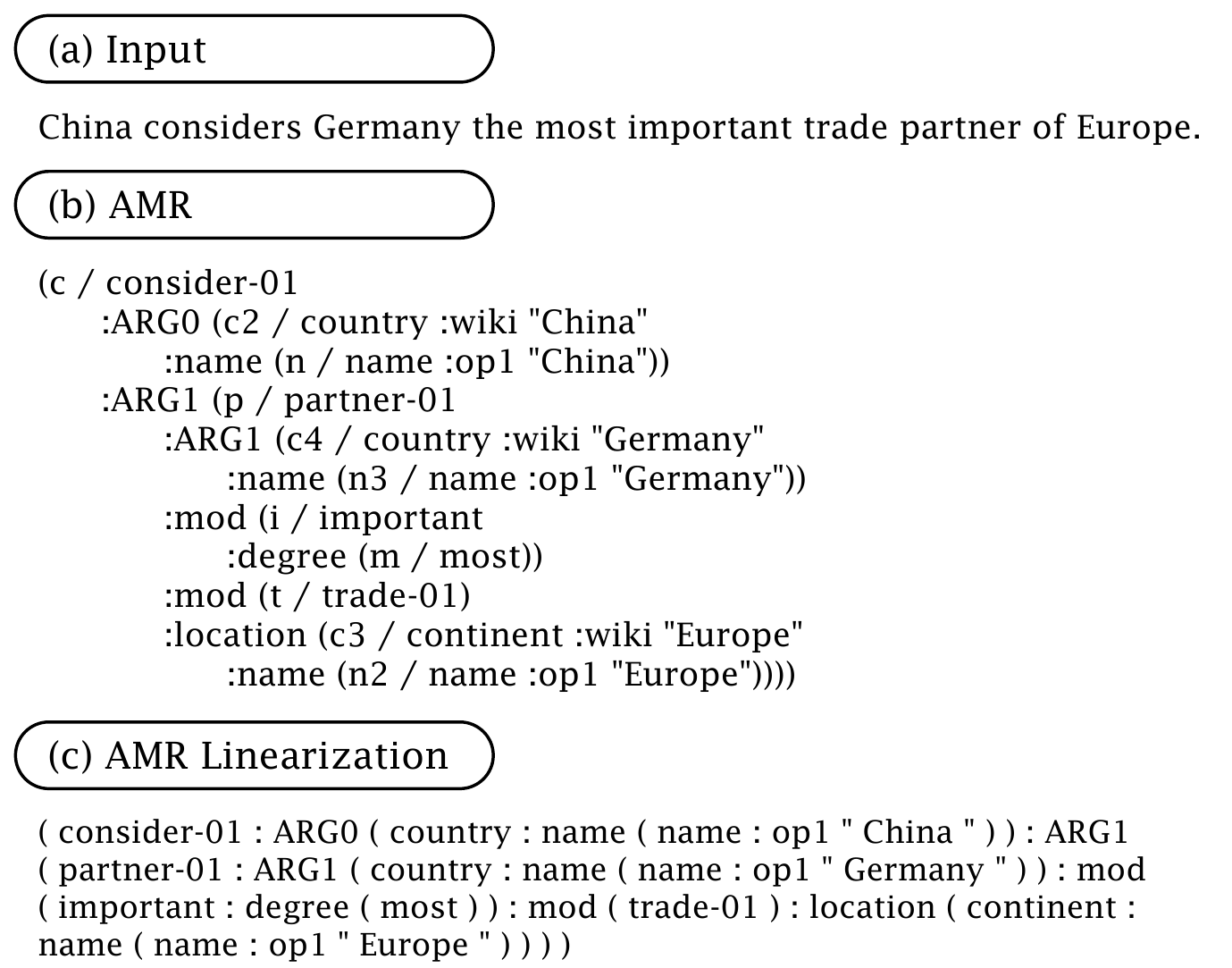}
\end{center}
\caption{An example of seq2seq-based AMR parsing.} \label{fig:amr_example}
\end{figure}

Similar to many NLP tasks, the performance of AMR parsing is much restricted by the size of human-curated dataset. For example, even recent AMR 2.0 contains only 36.5K training AMRs. To alleviate the effect of such restriction, a previous attempt is to utilize large-scale unlabeled sentences with self-training~\cite{konstas_etal_acl_2017}. Alternatively, a more recent feasible solution is to resort to pre-training which builds pre-trained models on large-scale (unlabeled) data. Linguistic knowledge captured in pre-trained models can then be properly incorporated into the training of an AMR parser. However, the widely used pre-trained models such as ELMO~\cite{peter_etal_naacl_2017} and BERT~\cite{devlin_etal_naacl_2019} may not work as expected for building a state-of-the-art seq2seq AMR parser. The reasons are two-fold. On the one hand, previous studies on both seq2seq-based AMR parsing and AMR-to-text generation demonstrate the necessity of a shared vocabulary for the source and target sides~\cite{ge_etal_ijcai_2019,zhu_etal_emnlp_2019}. Using pre-trained models like BERT as pre-trained encoders for AMR parsing, however, will violate the rule of sharing a vocabulary. On the other hand, pre-trained models such as BERT are basically tuned for the purpose of representing sentences instead of generating target sequences. According to~\citet{zhu_etal_iclr_2020}, by contrast to using BERT directly as the encoder, a more reasonable approach is to utilize BERT as an extra feature or view BERT as an extra encoder. See Section~\ref{sect:bert4amr} for more detailed discussions on the effect of BERT on AMR parsing. 

In this paper, we propose to pre-train seq2seq models that aim to capture different linguistic knowledge from input sentences\ignore{ and to properly generate output sequences}. To build such pre-trained models, we explore three different yet relevant seq2seq tasks, as listed in Table~\ref{tbl:tasks}. Here, machine translation acts as the most representative seq2seq task which takes a bilingual dataset as the training data. According to~\citet{shi_etal_emnlp_2016} and~\citet{li_etal_acl_2017}, a machine translation system with good performance requires the model to well derive linguistic information from input sentences. The other two tasks require auto-parsed syntactic parse trees and AMR graphs as the training data, respectively. It is worth noting that the pre-training task of AMR parsing is in the similar spirit of self-training~\cite{konstas_etal_acl_2017}.

\begin{table}[]
\setlength{\tabcolsep}{3pt}
    \small
    \centering
    \begin{tabular}{l|lll}
    \hline
         \bf Task & \bf Dataset & \bf Source & \bf Target  \\
    \hline
        machine translation & gold & sentence & sentence\\ 
        syntactic parsing  & silver & sentence & tree sequence \\ 
        AMR parsing  & silver & sentence & AMR sequence\\
    \hline
    \end{tabular}
    \caption{Three seq2seq learning tasks explored in this paper to obtain pre-trained models. Here \textit{silver} dataset indicates that the sequences in the target-side are generated automatically .}
    \label{tbl:tasks}
\end{table}

In order to investigate whether various seq2seq pre-trained models are complementary to each other in the sense that they can be learned jointly to achieve better performance, we further explore joint learning of several pre-training tasks and evaluate its effect on AMR parsing. In addition, motivated by~\citet{li_hoiem_pami_2018}, we extend the vanilla fine-tuning method to optimize for both the performance of AMR parsing and response preservation of the pre-trained models. Detailed experimentation on two widely used English benchmarks shows that our approach substantially improves the performance, which greatly advances the state-of-the-art. This is very encouraging since we achieve the state-of-the-art by simply making use of the generic seq2seq framework rather than designing sophisticated AMR parsing models.

\section{Baseline: AMR Parsing as Seq2Seq Learning}
\paragraph{Seq2Seq Modeling.} The encoder in the {\em Transformer}~\cite{vaswani_etal_nips_2017} consists of a stack of multiple identical layers, each of which has two sub-layers: one implements the multi-head self-attention mechanism and the other is a position-wise fully connected feed-forward network. The decoder is also composed of a stack of multiple identical layers. Each layer in the decoder consists of the same sub-layers as in the encoder layers plus an additional sub-layer that performs multi-head attention to the output of the encoder stack. See~\citet{vaswani_etal_nips_2017} for more details.

\paragraph{Pre-Processing: Linearize AMR Graph to Target Sequence.} As in~\citet{noord_bos_2017}, we obtain simplified AMRs by removing variables and wiki links. Variables in AMR graphs are only necessary to indicate co-referring nodes and they do not carry any semantic information by themselves. Therefore, AMR graphs are first converted into AMR trees by removing variables and duplicating the co-referring nodes. Then newlines present in an AMR tree are replaced by spaces to get a sequence. Figure~\ref{fig:amr_example}(c) illustrates the linearization result of the AMR graph in Figure~\ref{fig:amr_example}(b). Based on the data of sentences paired with linearized AMR graphs, we train a seq2seq model whose outputs are also linearized AMRs.

\paragraph{Post-Processing: Recover AMR Graph from Target Sequence.} The output from {\em Transformer} is an AMR sequence without variables, wiki-links, and co-occurrent variables. Moreover, the output may contain brackets that do not match, resulting incomplete concepts. To recover its full graph, the post-processing should restore information removed in pre-processing by assigning a unique variable to each concept, pruning duplicated and redundant material, performing Wikification, and restoring co-referring nodes. Meanwhile, it should fix incomplete concepts.

We use the pre-processing and post-processing scripts provided by~\citet{noord_bos_2017}.~\footnote{\url{ https://github.com/RikVN/AMR}}

\section{Seq2Seq Pre-training for AMR Parsing}
In this section, we first present our single pre-training approach, followed by the joint pre-training approach on two or more pre-training tasks. Then we present our fine-tuning methods.

\subsection{Single Pre-training}
To be consistent with the seq2seq model for AMR parsing, the pre-trained models in this paper are all built on the Transformer. That is, for each pre-training task listed in Table~\ref{tbl:tasks}, we learn a seq2seq model which will be used to initialize seq2seq model for AMR parsing in the fine-tuning phase. When building the pre-trained models, we merge all the source and target sides of the three pre-training tasks, and construct a shared vocabulary. Moreover, in all the models we share vocabulary embeddings for both the source and target sides.

\paragraph{PTM-MT} is a seq2seq neural machine translation (NMT) model which is trained on a publicly available bilingual dataset. According to findings in~\citet{goldberg_arxiv_2019} and \citet{jawahar_etal_acl_2019}, the Transformer encoder is strong in capturing syntax and semantics from source sentences, which is helpful to AMR parsing.

\begin{figure}[t]
\begin{center}
\includegraphics[width=2.5in]{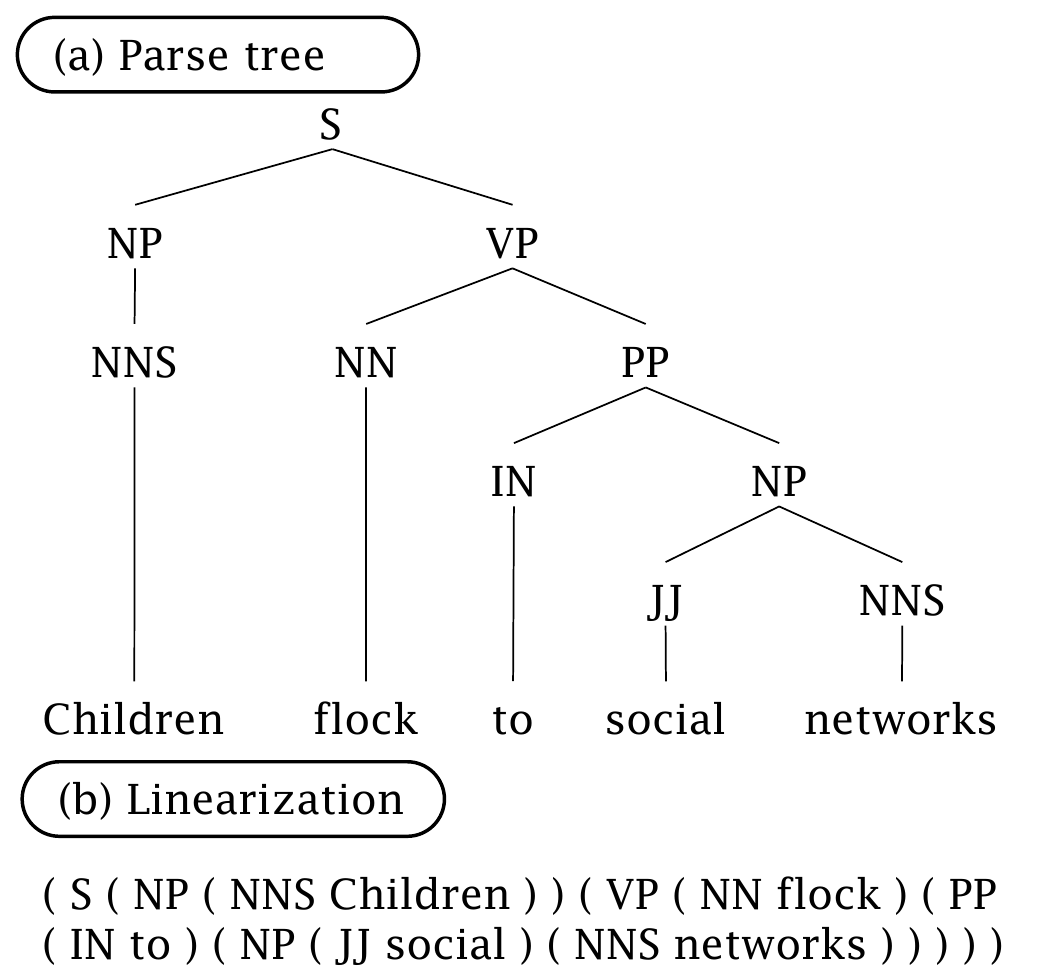}
\end{center}
\caption{A linearization example of the parse tree for the sentence of \textit{Children flock to social networks}.} \label{fig:syn_example}
\end{figure}

\paragraph{PTM-SynPar} is a seq2seq constituent parsing model. Building such a model requires a training dataset which consists of sentences paired with constituency parse trees. To construct a silver treebank, we parse the English sentences in the bilingual data for MT by using an off-the-shelf parser\ignore{ to obtain a silver treebank}. Then we linearize the automatic parse trees to get syntax sequences, as illustrated in Figure~\ref{fig:syn_example}. Note that in the linearization, we let the output contain the words from the source sentence. The motivation here is to regard parsing as a language generation problem, similar to the idea in~\citet{choe_charniak_emnlp_2016}. 

\paragraph{PTM-SemPar} is a seq2seq AMR parsing model trained on a silver corpus of auto-parsed AMR graphs. To construct such a corpus, we apply the baseline system of AMR parsing to process the English sentences in the bilingual MT corpus. Then we adopt the linearization process illustrated in Figure~\ref{fig:amr_example} to obtain source-target pairs. Finally, we train a seq2seq-based AMR parsing model on the silver corpus that will be used as a pre-trained model.

\subsection{Joint Pre-training}
\begin{figure}[t]
\begin{center}
\includegraphics[width=2.8in]{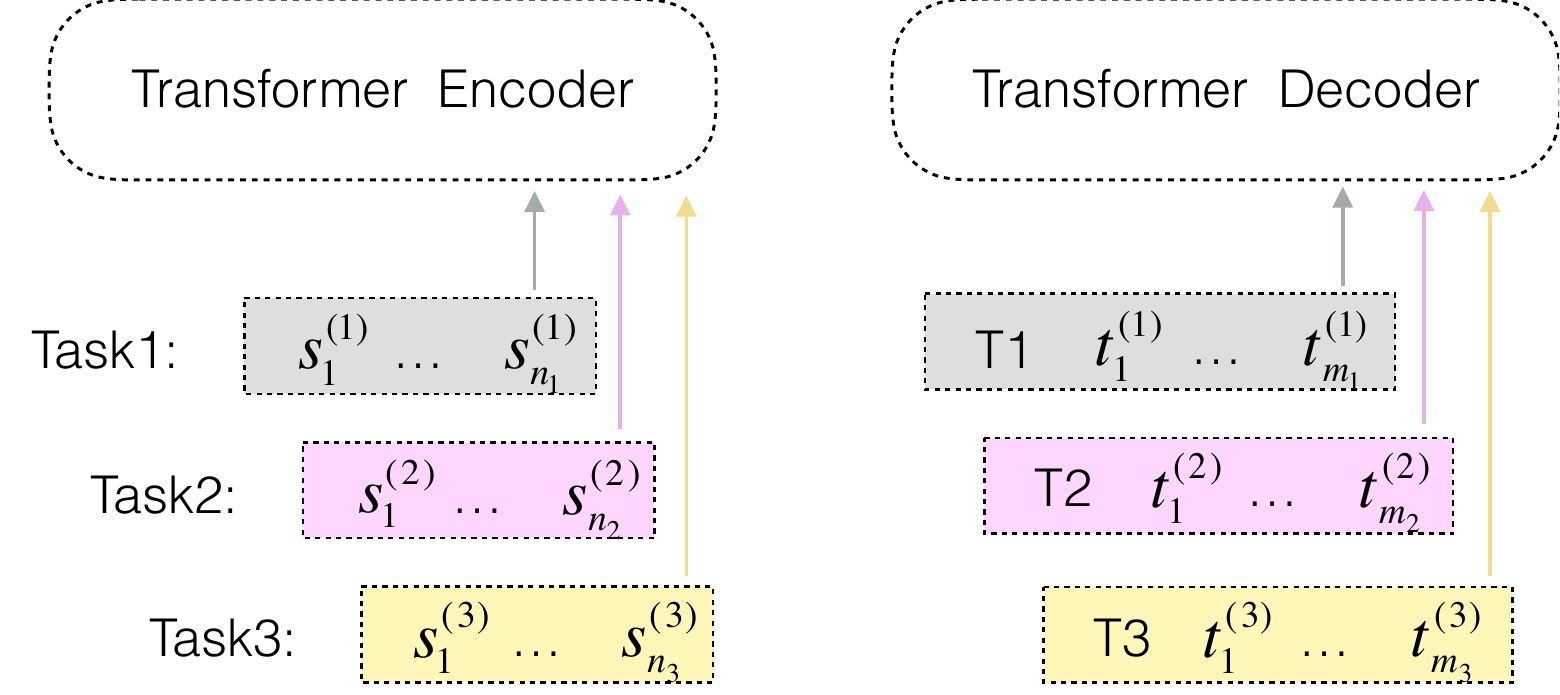}
\end{center}
\caption{Illustration of the joint pre-training approach.} \label{fig:joint_training}
\end{figure}

Intuitively, the above described single pre-trained models can capture linguistic features from different perspectives. One question is whether these models are complementary when they are properly used to initialize a seq2seq-based AMR parser. To empirically answer this question, we propose to build pre-trained models through jointly learning multiple pre-training tasks. Inspired by the zero-shot approach proposed for multi-lingual neural machine translation~\cite{johnson_etal_tacl_2017}, we add a unique preceding tag to the target side of training data to distinguish the task of each training instance, as illustrated in Figure~\ref{fig:joint_training}.

With such tagged training instances, multi-task learning is actually quite straightforward. We simply combine the training data of all the pre-training tasks that we are focusing on and then feed the combined training data to the Transformer model. The training process interleaves training data from each task. For example, we update parameters on a batch of training instances from task1 and then update parameters on a batch of training instances from task2, and the process iterates. With such a joint training strategy, we obtain four joint pre-trained models, i.e., PTM-MT-SynPar, PTM-MT-SemPar, PTM-SynPar-SemPar, and PTM-MT-SynPar-SemPar. Names of the models can tell what pre-training tasks are learned jointly.

\subsection{Fine-tuning Methods}
Given a pre-trained model, we can directly fine-tune it on a gold AMR corpus to train an AMR parser. For this purpose we use two different fine-tuning methods. In the following we first present the vanilla fine-tuning method, and then extend it under the framework of multi-task learning. For simplicity, we refer to the latter method as Multi-Task Learning (MTL) fine-tuning hereafter.

\paragraph{Vanilla Fine-Tuning} optimizes the parameters of an existing pre-trained seq2seq models to train AMR parsing on a gold AMR corpus. Fine-tuning adapts the shared parameters to make them more discriminative for AMR parsing, and the low learning rate is an indirect mechanism to preserve some of the representational structure captured in the pre-training models. 

\paragraph{MTL Fine-Tuning} is designed to attack the potential drawback of the vanilla fine-tuning method. In vanilla fine-tuning, optimizing model parameters to train AMR parsing presents a potential risk of overfitting. Inspired by~\citet{li_hoiem_pami_2018}, we propose to optimize for high accuracy of AMR parsing while preserving the performance on the pre-training tasks. Preservation of the performance on the pre-training tasks can be regarded as a regularizer for the training of AMR parsing. To implement such MTL fine-tuning, we once again adopt the generic multi-task learning framework depicted in Figure~\ref{fig:joint_training}. 

Now the question left behind is how to obtain fine-tuning instances for pre-training tasks.  To this end, we use the pre-trained model focused and input sentences of gold AMR corpus to generate fine-tuning instances for pre-training tasks. Formally speaking, given an instance $\{s,t^{(0)}\}$ of the fine-tuning task , and a pre-trained model learned from $k$ pre-training tasks, we first feed the pre-trained model with input $s$ and obtain its $k$ outputs, i.e. $t^{1},\cdots,t^{k}$ for the $k$ pre-training tasks, respectively. Therefore, each input $s$ in the fine-tuning task is now equipped with $k+1$ outputs, one for the fine-tuning task while the other $k$ for the $k$ pre-training tasks. Meanwhile, each output is associated with a unique preceding tag which indicates the corresponding task.

Please also note that we do not apply MTL fine-tuning to the pre-training task of AMR parsing. This is because the fine-tuning task is the same as the pre-training task. For example, for the pre-trained model PTM-MT-SynPar-SemPar, in MTL fine-tuning we only keep the pre-training tasks of MT and syntactic parsing.

\section{Experimentation}
In this section, we report the performance of our seq2seq pre-training approach to AMR parsing.

\subsection{Experimental Settings}
\label{sect:setting}
\paragraph{Pre-training Dataset and Pre-trained Models} 

For pre-trained models, we use the WMT14 English-to-German dataset\footnote{\url{https://www.statmt.org/wmt14/translation-task.html}} which consists of about 3.9M training sentence pairs after filtering out long and imbalanced pairs. To obtain syntactic parse trees for the source sentences, we utilize toolkit AllenNLP~\citep{gardner_etal_acl_2017} which is trained on Penn Treebank~\cite{marcus_etal_cl_1993}. To obtain AMR graphs for the source sentences, we utilize our baseline AMR parsing system. Then we merge English/German sentences and linearized parse trees, and AMR graphs together and segment all the tokens into subwords by byte pair encoding (BPE)~\cite{sennrich_etal_acl_2016} with 20K operations. 

We implement above pre-trained models based on \textit{OpenNMT-py}~\cite{klein_etal_acl_2017}.\footnote{\url{https://github.com/OpenNMT/OpenNMT-py}} For simplicity, we unify parameters of these models as the Transformer-base model in~\citet{vaswani_etal_nips_2017}. The number of layers in encoder and decoder is 6 while the number of heads\ignore{ for transformer attention} is 8. Both the embedding size and the hidden size are 512 while the size of feedforward network is 2048. Moreover, we use Adam optimizer~\cite{kingma_ba_iclr_2015} with $\beta_1$ of 0.9 and $\beta_2$ of 0.998. Warm\_up step, learning rate, dropout rate and label smoothing epsilon are 16000, 2.0, 0.1 and 0.1 respectively. In addition, we set the batch token-size to 8,192. We train the models for 300K steps and choose the model with the best performance on WMT2014 English-to-German development set as the final pre-trained model.

\paragraph{AMR Parsing Benchmarks}
We evaluate AMR performance on AMR 1.0 (LDC2015E86) and AMR 2.0 (LDC2017T10). The two datasets contain 16,833 and 36,521 training AMRs, respectively, and share 1,368 development AMRs and 1,371 testing AMRs. All the source sentences and linearized AMRs are segmented into subwords by using the BPE trained for the pre-trained models. 

To fine-tune the pre-trained models for AMR parsing, we follow the settings of hyper-parameters used for training pre-trained models.

\paragraph{Evaluation Metrics}
For evaluation purpose, we use the AMR-evaluation toolkit to evaluate parsing performance in Smatch and other fine-grained metrics~\cite{cai_knight_eacl_2013,damonte_etal_eacl_2017}. We report results of single models that are tuned on the development set.

\subsection{Experimental Results}
Table~\ref{tbl:performance} presents the comparison of our approach and related studies on the test sets of AMR 1.0 and AMR 2.0. From the results, we have the following observations:

\begin{itemize}
    \item Pre-trained models on a single task (i.e., from \#2 to \#6) significantly improve the performance of AMR parsing, indicating seq2seq pre-training is helpful for seq2seq-based AMR parsing. We also note that the pre-trained model of NMT achieves the best performance, followed by the pre-trained models on AMR parsing and on syntactic parsing. This indicates that seq2seq AMR parsing benefits more from pre-training tasks that require the encoder be able to capture the semantics from source sentences. 
    \item Joint pre-trained models on two or more pre-training tasks further improve the performance of AMR parsing. However, in the presence of NMT pre-training task, the benefits from joint pre-training with either AMR parsing, syntactic parsing or both are shrunk. 
    \item MTL fine-tuning consistently outperforms the vanilla fine-tuning method. For example, on single pre-training tasks, MTL outperforms vanilla fine-tuning by 1.5 $\sim$ 2.0 Smatch F1 scores while on joint pre-training tasks, the improvements of MTL over vanilla fine-tuning instead decrease.
    \item With twice training sentences in AMR 2.0, overall the performance on AMR 2.0 is higher than that on AMR 1.0. However, the gap between the performance on AMR 2.0 and AMR 1.0 gets smaller when we move from single pre-training models to joint pre-training models. For example, based on PTM-MT-SynPar-SemPar, the performance gap is 1.1 in Smatch F1 scores, much less than the performance gap 6.9 between their corresponding baselines.
    \item Finally, our approach achieves the best reported performance on AMR 1.0 and the performance on AMR 2.0 is higher than or close to that achieved by previous studies which use BERT. This is very encouraging taking into consideration the fact that our seq2seq model is much simper than the graph-based models proposed in related studies~\cite{zhang_etal_acl_2019,zhang_etal_emnlp_2019,naseem_etal_acl_2019, cai_lam_acl_2020}.
\end{itemize}

\begin{table*}[]
   \small
   \setlength{\tabcolsep}{4pt} 
    \centering
    \begin{tabular}{l|r|l|ccc|ccc}
        \hline
         \multirow{2}{*}{\bf \#} & \multirow{2}{*}{\bf Pre-trained Model} &  \multirow{2}{*}{\bf Fine-Tune} & \multicolumn{3}{c|}{\bf AMR 1.0} & \multicolumn{3}{c}{\bf AMR 2.0} \\
         \cline{4-9}
        & & & \bf P. & \bf R. & \bf F1 & \bf P. & R. & \bf F1\\
        \hline
        1 & None & None & 69.8 & 60.2 & 64.6 & 75.8 & 67.7 & 71.5 \\
        \hline
        \hline
        2 & \multirow{2}{*}{PTM-MT} & Vanilla & 78.8 & 69.5 & 73.8 & 80.0 & 74.3 & 77.1 \\
        3 & & MTL & 81.1 & 72.2 & 76.4 & 81.3 & 77.1 & 79.1 \\
        \hline
        4 & \multirow{2}{*}{PTM-SynPar} & Vanilla & 74.3 & 65.8 & 69.8 & 76.2 & 71.5 & 73.8 \\
        5 & & MTL & 76.7 & 68.1 & 72.2 & 78.0 & 72.8 & 75.3 \\
        \hline
        6 & PTM-SemPar & Vanilla  & 80.8 & 73.5 & 77.0 & 80.8 & 75.2 & 77.9 \\
        \hline
        \hline
        7 & \multirow{2}{*}{\shortstack[l]{PTM-MT-SynPar}} & Vanilla & 79.1 & 70.5 & 74.6 & 79.5 & 75.0 & 77.1 \\
        8 & & MTL & 81.2 & 74.0 & 77.5 & 81.5 & 77.6 & 79.5 \\
        \hline
        9 & \multirow{2}{*}{\shortstack[l]{PTM-MT-SemPar}} & Vanilla & 82.3 & 75.4 & 78.7 & \bf 82.4 & 77.3 & 79.7 \\
        10 & & MTL & 82.4 & 74.6 & 78.3 & 82.3 & 78.0 & 80.1 \\
        \hline
        11 & \multirow{2}{*}{\shortstack[l]{PTM-SynPar-SemPar}} & Vanilla & 81.6 & 74.0 & 77.6 & 81.1 & 76.3 & 78.6 \\
        12 & & MTL &81.8 & 74.0 & 77.7 & 81.3 & 76.8 & 79.0 \\
        \hline
        \hline
        13 & \multirow{2}{*}{\shortstack[l]{PTM-MT-SynPar-SemPar}} & Vanilla & 82.4 & 75.4 & 78.7 & 82.1 & 77.6 & 79.8 \\
        14 & & MTL & \bf 82.6 & \bf 75.9 & \bf 79.1 & 82.3 & \bf 78.3 & \bf 80.2 \\
        \hline
        \hline
        \multicolumn{9}{c}{\bf Previous work without extra resources}\\
        \hline
        \multicolumn{3}{r|}{Graph Prediction\small{\citep{lyu_titov_acl_2018}} } & - & - & - & - & - & 74.4\\
        \multicolumn{3}{r|}{Prediction\small{\citep{guo_lu_emnlp_2018}}} & - & - & - & - & - & 69.8\\
        \multicolumn{3}{r|}{Prediction\small{\citep{groschwitz_etal_acl_2018}}} & - & - & - & - & - & 71.0\\
        \multicolumn{3}{r|}{Seq2Seq\small{\citep{ge_etal_ijcai_2019}}} & - & - & - & 74.0 & 68.1  & 70.9\\
        \multicolumn{3}{r|}{Seq2Seq\small{\citep{cai_lam_emnlp_2019}}} & - & - & - & - & -  & 73.2\\
        \multicolumn{3}{r|}{Graph\small{\citep{cai_lam_acl_2020}}} & - & - & 71.2 & - & -  & 77.3\\
        \hline
        \hline
        \multicolumn{9}{c}{\bf Previous work with extra resources}\\
        \hline
        \multicolumn{3}{r|}{Seq2Graph\small{\citep{zhang_etal_acl_2019}}$\dagger$} & - & - & 70.2 & - & - & 76.3\\
        \multicolumn{3}{r|}{Seq2Graph\small{\citep{zhang_etal_emnlp_2019}}$\dagger$} & - & - & 71.3 & - & - & 77.0\\
        \multicolumn{3}{r|}{RL\small{\citep{naseem_etal_acl_2019}}$\dagger$} & - & - & - & - & - & 75.5\\
        \multicolumn{3}{r|}{Seq2Seq\small{\citep{ge_etal_ijcai_2019}}$\ast$} & - & - & - & 77.7 & 71.1  & 74.3\\
        \multicolumn{3}{r|}{Graph\small{\citep{cai_lam_acl_2020}}$\dagger$} & - & - & 75.4 & - & -  & \bf 80.2\\
        \hline
    \end{tabular}
    \caption{Smatch scores on the test sets of AMR 1.0 and AMR 2.0. $\dagger$ is for using BERT as extra resource while $\ast$ for using other resources.}
    \label{tbl:performance}
\end{table*}

Table~\ref{tbl:detailed_comparison} compares the performance of our best system and the systems reported recently with fine-grained metrics. We obtain the best performance for Reentrancies, NER, and SRL. Compared to the systems of Z'19a, Z'19b, and C'20, we achieve lower performance for Wiki and Negations. One possible reason for our relatively lower performance on Wiki and Negations is that unlike above three systems, in this paper we do not anonymize named entities and do not use an extra algorithm to add polarity attributes.

\begin{table}[th]
    \setlength{\tabcolsep}{2pt}
    \centering
    \small
    \begin{tabular}{l|cccccc|c}
    \hline
    \bf{Metric} & \bf{C'19} & \bf{G'19} & \bf{N'19} & \bf{Z'19a} & \bf{Z'19b} & \bf{C'20} & \bf{Our} \\
    \hline
    Smatch       & 73.2 & 74.3 & 75.5 & 76.3 & 77 & \bf{80.2} & \bf{80.2} \\
    \hline
    Unlabeled    & 77.0 & 77.3 & 80   & 79.0 & 80 & 82.8      & \bf{83.7} \\
    No WSD       & 74.2 & 74.8 & 76   & 76.8 & 78 & \bf{80.8} & \bf{80.8} \\
    Reentrancy   & 55.3 & 58.3 & 56   & 60.0 & 61 & 64.6      & \bf{66.5} \\
    Concepts     & 84.4 & 84.2 & 86   & 84.8 & 86 & \bf{88.1} & 87.4 \\
    NER          & 82.0 & 82.4 & 83   & 77.9 & 79 & 81.1      & \bf{85.4} \\
    Wiki         & 73.2 & 71.3 & 80   & 85.8 & 86 & \bf{86.3} & 75.1 \\
    Negations    & 62.9 & 64.0 & 67   & 75.2 & 77 & \bf{78.9} & 71.5 \\
    SRL          & 66.7 & 70.4 & 72   & 69.7 & 71 & 74.2      & \bf{78.9} \\
    \hline
    \end{tabular}
    \caption{Detailed F1 scores on AMR 2.0 test set. Here, C'19 is for \citet{cai_lam_emnlp_2019}, G'19 for \citet{ge_etal_ijcai_2019}, N'19 for \citet{naseem_etal_acl_2019}, Z'19 for \citet{zhang_etal_acl_2019}, Z'19b for \citet{zhang_etal_emnlp_2019}, C'20 for \citet{cai_lam_acl_2020}
    }
    \label{tbl:detailed_comparison}
\end{table}

\section{Analysis and Discussion}
\label{sect:analsysi}
In this section, we conduct more analysis on AMR parsing with pre-trained models. In the following all the results are obtained on AMR 2.0.

\subsection{Effect of BERT on Seq2Seq AMR Parsing}
\label{sect:bert4amr}
To explore the effect of BERT on seq2seq AMR parsing, motivated by~\citet{zhu_etal_iclr_2020}, we use BERT in various ways to boost the performance of AMR parsing. 

\begin{figure}[t]
\begin{center}
\includegraphics[width=3.0in]{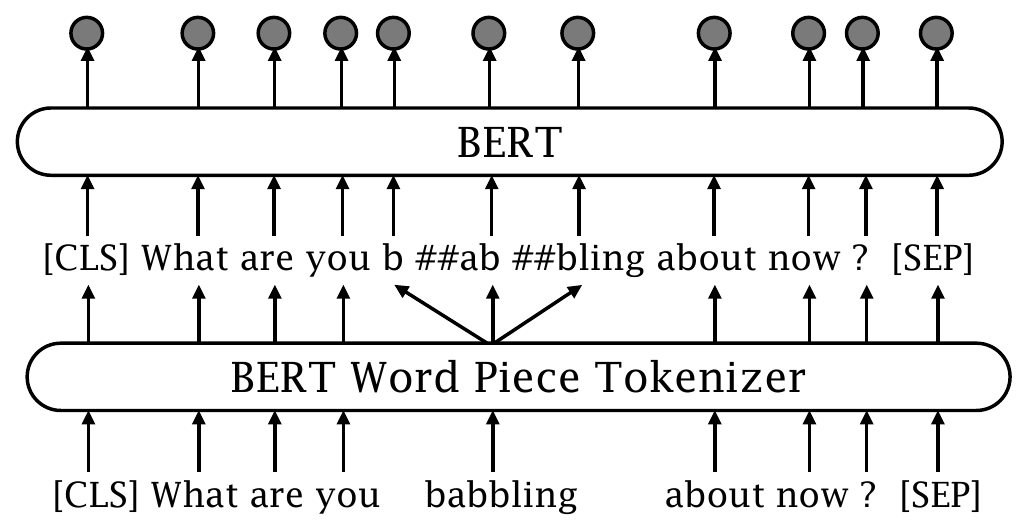}
\end{center}
\caption{Illustration of obtaining BERT hidden states for an given sentence.} 
\label{fig:bert}
\end{figure}

Given an input sentence $x=\left(x_1, \cdots,x_n\right)$ with $n$ words, the BERT tokenizer segments it into a subword sequence $x'=\left(x'_1, \cdots,x'_m\right)$ with $m$ subwords. Then BERT returns a hidden state sequence $b=\left(b_1,\cdots,b_m\right)$ in shape $\mathbb{R}^{m\times d_{BERT}}$, where $d_{BERT}$ is the size of BERT hidden states (e.g., $d_{BERT}$=768 in our experiment). Figure~\ref{fig:bert} illustrates the process of obtaining BERT hidden states for an input sentence. Next we use the following methods to properly incorporate BERT hidden states $b$ into Transformer-based AMR parsing.

\begin{itemize}
\item BERT as embedding, which uses $f\left(bW^B\right)$ as input of the the Transformer encoder, where $W^B \in \mathbb{R}^{d_{BERT}\times d}$ are model parameters to be learned, $d$ is the model size for seq2seq AMR parsing, and $f$ is the activation function \textit{ReLu}.
\item BERT as encoder, which uses $f\left(bW^B\right)$ as the output of the Transformer encoder. That is to say, we replace the Transformer encoder with BERT. 
\item BERT as extra feature, which views $b$ as extra features for an input sentence $x'$. The input of the Transformer encoder is defined as $f\left([b, \left(Emb\left(x'\right) + Pos\left(x'\right)\right)]W^E\right)$, where $[\cdot,\cdot]$ represents the operation of concatenation, $Emb\left(x'\right)$ and $Pos\left(x'\right)$ return the word embeddings and position embeddings of $x'$ respectively, and $W^E\in\mathbb{R}^{(d+d_{BERT})\times d}$ are model parameters to be learned.
\item BERT as extra encoder, which adds a sub-layer, i.e, BERT-context-attention sub-layer, in the Transformer decoder after the masked-self-attention sub-layer and the context-attention sub-layer. The BERT-context-attention sub-layer works in a similar way as the context-attention sub-layer by attending to BERT hidden states $f\left(bW^B\right)$.
\end{itemize}

Meanwhile, we also provide another Transformer-based baseline in which we segment input sentences into subwords with the BERT tokenizer. For all above experiments, the source-side vocabulary is the set of subwords in training sentences segmented by the BERT tokenizer while the target-side vocabulary is the set of subwords in training AMRs segmented by BPE mentioned in Section~\ref{sect:setting}.  

Table~\ref{tbl:bert} compares the performance of AMR  parsing when incorporating BERT in various methods. By comparing the performance of \#1 in Table~\ref{tbl:bert} against the baseline \#1 in Table~\ref{tbl:performance}, we observe a drop of Smatch F1 score from 71.5 to 70.0, indicating that it is important to share vocabulary for seq2seq AMR parsing. Based on the baseline of not sharing vocabulary, the four different methods of incorporating BERT result in very different performance ranging from 71.5 to 75.2 in Smatch F1 score. Among them, incorporating BERT as embedding or extra feature\ignore{, as recent studies~\cite{zhang_etal_acl_2019,cai_lam_acl_2020} do,} achieves similar performance, which is much higher than the performance of incorporating BERT as either encoder or extra encoder. This suggests that rather than straightly feeding BERT hidden states into a decoder, it is important to feed them into an encoder first. However, our pre-trained seq2seq models, even on a single pre-training task (i.e., \#3, \#5, \#6) outperform using BERT, indicating the effectiveness of pre-trained seq2seq models for AMR parsing.

\begin{table}[t]
    \centering
    \begin{tabular}{l|l|ccc}
    \hline 
    \bf \# & \bf Methods & \bf P. & \bf R. & \bf F1\\
    \hline
        1 & None &  73.5 & 66.9 & 70.0 \\
         \hline
        2 & BERT as embedding &  78.1 & 72.2 & 75.1\\
        3 & BERT as encoder&  75.5 & 68.0 & 71.5 \\
        4 & BERT as extra feature&  79.2 & 71.5 & 75.2 \\
        5 & BERT as extra encoder &  75.1 & 68.2 & 71.5 \\
         \hline
    \end{tabular}
    \caption{Smatch scores on AMR 2.0 when incorporate BERT in various methods.}
    \label{tbl:bert}
\end{table}

\subsection{Effect of Training Data Sizes on Pre-training Models}
In this section we investigate the impact of the size of pre-training data to check whether AMR parsing benefits more from pre-trained models that are trained on larger datasets. To this end, we randomly use 20\%, 40\%, 60\%, and 80\% of the full pre-training instances to train the pre-trained models, respectively. 

\begin{figure}[]
\begin{center}
\includegraphics[width=3.0in]{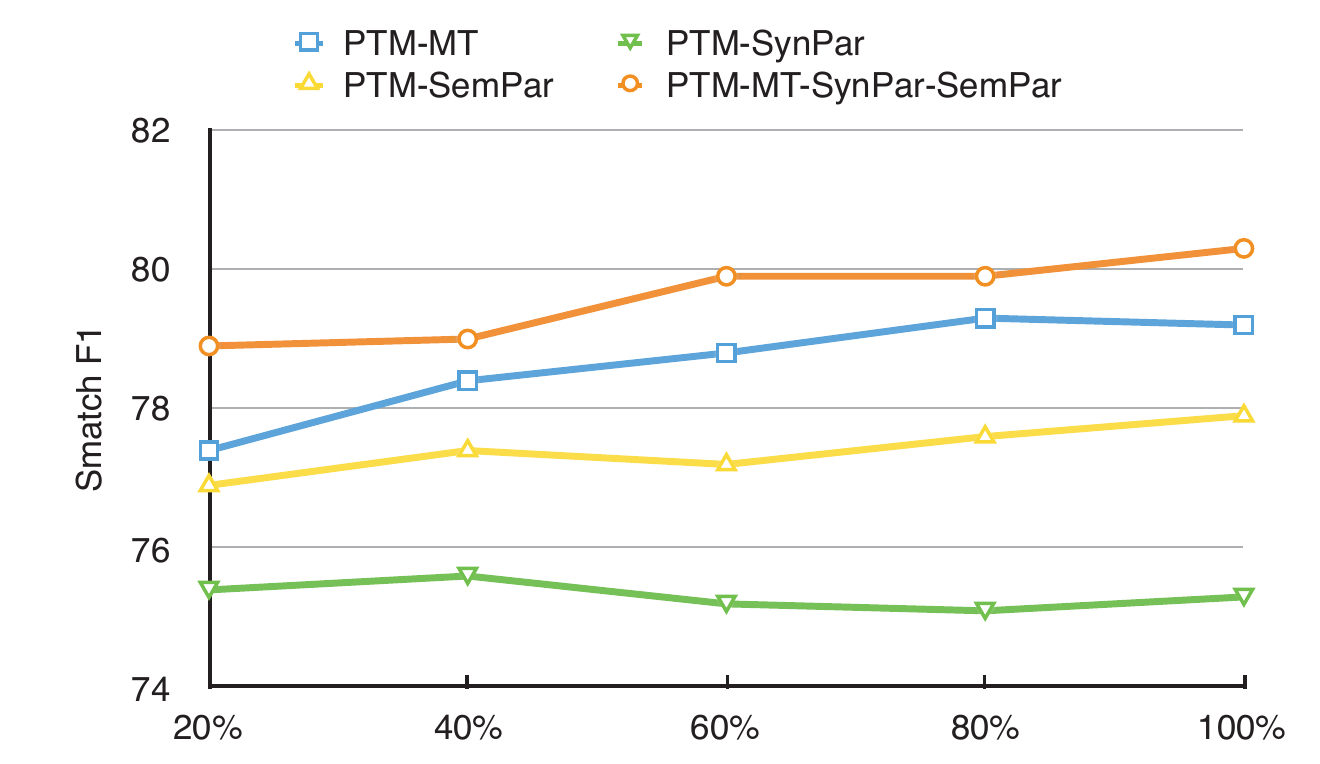}
\end{center}
\caption{Learning curve over the number of training sentences in pre-training datasets.} 
\label{fig:curve}
\end{figure}

As shown in Figure~\ref{fig:curve}, except syntactic parsing (i.e., PTM-SynPar), the pre-training models on the other three kinds of pre-training tasks achieve higher AMR parsing performance with the increasing of training data sizes. Based on the learning curve, we suspect there still exists much room for further improvements if we enlarge the training data of pre-training tasks.


\ignore{
\subsection{Effect of Sentence Lengths}

\begin{figure}[]
\begin{center}
\includegraphics[width=3.0in]{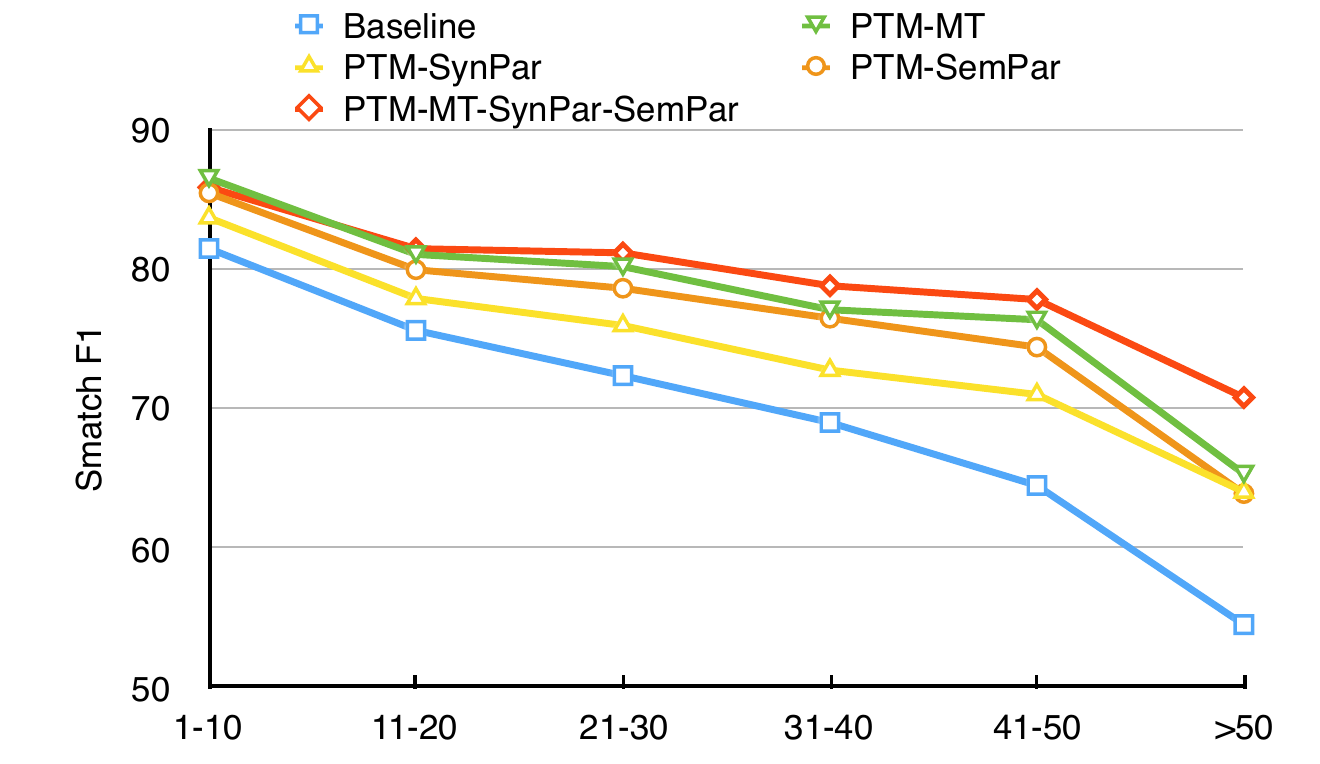}
\end{center}
\caption{AMR Parsing performance with respect to the lengths of the sentences.} 
\label{fig:sent_len}
\end{figure}

In order to explore the effect of sentence lengths on AMR parsing performance, we divide the test set into 6 groups according to the sentence lengths and evaluate
their performance on five different pre-trained seq2seq models respectively. As shown in Figure~\ref{fig:sent_len}, our pre-trained models significantly outperform the baseline over all sentence lengths. We also notice that the performance of models gradually decreases when the sentence length becomes longer. We think that AMR graphs of long sentences are more likely to have entities and reentrancies, which makes the parsing more challenging.
}

\subsection{Effect of Different Pre-Training Components on Seq2Seq AMR Parsing}
When adapt a pre-trained model to AMR parsing, we initialize the whole seq2seq Transformer model of AMR parsing with the counterpart of the pre-trained model. However, it is unveiled what part of initialization contributes most. To this end, we decompose the whole seq2seq model into three components, i.e., (shared) word embedding, encoder and decoder. The three components take account of 31.1\%, 29.5\% and 39.4\%  of parameters, respectively. Then we do ablation study by accumulating the initialization using the pre-trained model while the other components will be randomly initialized. 

We use the PTM-MT-SynPar-SemPar pre-trained model as representative (i.e., \#14 in Table~\ref{tbl:performance}). Table~\ref{tbl:ablation} presents the performance. From the table, we observe that with well-learned word embedding, we substantially boost the performance from 71.5 in Smatch F1 score to 78.4 while initializing the other two components with the pre-trained model leads to another 1.8 improvement in Smatch F1 score (i.e., from 78.4 to 80.2). 

\begin{table}[t]
    \centering
    \small
    \begin{tabular}{l|ccc}
    \hline
       \bf Pre-trained Initialization  & \bf P. & \bf R. & \bf F1 \\
       \hline
       None & 75.8 & 67.7 & 71.5 \\
       \hline
       Embedding & 80.7 & 76.3 & 78.4\\
       Embedding + Encoder & 81.3 & 77.2 & 79.2 \\
       Embedding + Decoder & 80.7 & 76.5 & 78.5 \\
       \hline
       All  & \bf 82.3 & \bf 78.3 & \bf 80.2\\
        \hline
    \end{tabular}
    \caption{Smatch F1 scores on the test sets of AMR2.0 when initialize different components of seq2seq model with a pre-trained model. Here we use MTL as fine-tuning method.}
    \label{tbl:ablation}
\end{table}

\subsection{Effect of Pre-trained Models Trained on Different Datasets}
As shown in Table~\ref{tbl:performance}, the pre-trained model of PTM-SynPar (or PTM-SemPar) significantly improves the performance AMR parsing from 71.5 to 75.3 (or 77.9) in Smatch F1 score. However, in the presence of PTM-MT, joint pre-training with either PTM-SynPar, PTM-SemPar, or both gives another up to 1.0 improvement, suggesting that complementarity among the pre-trained models does exist but is relatively limited. We suspect that the overlapping is mainly due to the fact that we pre-train these models on the same source-side dataset. We conjecture that more improvement is potentially reachable if the pre-training tasks are trained on different datasets.  

To test the conjecture, we construct another silver dataset for both syntactic parsing and AMR parsing that is in the same size (i.e., 3.9M) as before. This is done by randomly selecting 3.9M English sentences from WMT14 English monolingual language model training data.\footnote{\url{http://statmt.org/wmt14/training-monolingual-news-crawl/news.2008.en.shuffled.gz}} Table~\ref{tbl:change_dataset} compares the Smatch F1 scores. From it, we observe consistent improvement if the pre-trained models are jointly trained on different datasets. For example, by replacing the pre-training dataset of AMR parsing with the new constructed dataset, we improve AMR parsing from 80.1 in Smatch F1 score to 81.4. This suggests that assigning different pre-training tasks with different datasets improves the performance of AMR parsing.


\subsection{Effect of Different Bilingual Datasets}
For the pre-training task of machine translation, we have chosen English-to-German (EN-DE) with 3.9M sentence pairs. However, it is still unclear whether it is critical to choose the right language pair. To this end, we move to WMT14 Englilsh-to-French (EN-FR) translation and randomly select 3.9M sentence pairs from its training dataset, as the same size of EN-DE translation. Table~\ref{tbl:change_nmt_dataset} compares the Smatch F1 scores when the pre-trained models are trained on different bilingual datasets. From it, we observe that pre-training on EN-FR dataset achieves even slight higher performance than that on EN-DE dataset. This further confirms our finding that AMR parsing can greatly benefit from machine translation.


\begin{table}[]
    \centering
    \small
    \begin{tabular}{l|l|c}
    \hline
      \# & \bf Pre-trained Model  &  \bf F1\\
    \hline
        1 & PTM-MT (WMT14B) & 79.1\\
    \hline
    \hline
        2 & \shortstack[l]{PTM-MT(WMT14B)-SemPar(WMT14B)} & 80.1 \\
    \hline
        3 & \shortstack[l]{PTM-MT(WMT14B)-SemPar(WMT14M)} & \bf 81.4 \\
    \hline
    \hline
        4 & \shortstack[l]{PTM-MT(WMT14B)-SynPar(WMT14B)} & 79.5 \\
    \hline
        5 & \shortstack[l]{\shortstack[l]{PTM-MT(WMT14B)-SynPar(WMT14M)}} & 79.9 \\
    \hline
    \end{tabular}
    \caption{Smatch F1 scores on the test set of AMR 2.0 when the pre-training tasks are trained on different datasets. Here WMT14B is for WMT14 English-to-German dataset while WMT14M is for WMT14 English monolingual dataset.}
    \label{tbl:change_dataset}
\end{table}

\begin{table}[]
\centering
\small
\begin{tabular}{l|l|l|l}
\hline
\textbf{\#} & \multicolumn{2}{l|}{\textbf{Pre-trained Model}} & \textbf{F1}   \\ 
\hline
1 & \multirow{2}{*}{PTM-MT on EN-DE} & Vanilla & 77.1 \\
2 &  & MTL & 79.1 \\ 
\hline 
3 & \multicolumn{1}{c|}{\multirow{2}{*}{PTM-MT on EN-FR}} & Vanilla & 77.5 \\
4 & \multicolumn{1}{c|}{} & MTL & \textbf{79.4} \\
\hline
\end{tabular}
\caption{Smatch F1 scores on the test set of AMR 2.0 when the pre-training tasks are trained on different bilingual dataset.}
\label{tbl:change_nmt_dataset}
\end{table}

\section{Related Work}
\label{sect:related_work}
We describe related work from two perspectives: pre-training and AMR parsing. 

\paragraph{Pre-training.} Pre-training a universal model and then fine-tuning the model on a downstream task have recently become a popular strategy in the field of natural language processing. Previous works on pre-training can be roughly grouped into two categories. One category of approaches is to learn static word embeddings such as word2vec~\cite{mikolov_etal_nips_2013} and GloVe~\cite{pennington_etal_emnlp_2014} while the other group builds dynamic pre-trained models that would also be used in downstream tasks. Representative examples in the latter group include~\citet{dai_le_nips_2015}, CoVe~\cite{mccann_etal_nips_2017}, ELMo~\cite{peter_etal_naacl_2017,edunov_etal_naacl_2019}, OpenAI GPT~\cite{radford_etal_note_2018}, and BERT~\cite{devlin_etal_naacl_2019}. Besides the aforementioned encoder-only (e.g., BERT) or decoder-only (e.g., GPT) pre-training approaches, recent studies also propose approaches to pre-training seq2seq models, such as MASS~\cite{song_etal_icml_2019}, PoDA~\cite{wang_etal_emnlp_2019}, PEGASUS~\cite{zhang_etal_icml_2020}, BART~\cite{lewis_etal_acl_2020}, and T5~\cite{raffel_etal_jml_2020}. 

\paragraph{AMR Parsing.} As a semantic parsing task that translates texts into AMR graphs, AMR parsing has received much attention in recent years. Diverse approaches have been applied to the task. ~\citet{flanigan_etal_acl_2014} pioneer the research work on AMR parsing by using a a two-stage approach: node identification followed by relation recognition. ~\citet{werling_etal_acl_2015} improve the first stage in the parser of~\citet{flanigan_etal_acl_2014} by generating subgraph aligned to lexical items. To avoid conducting AMR parsing from scratch, \citet{wang_etal_naacl_2015} propose to obtain AMR graphs from dependency trees by using a transition-based method. ~\citet{wang_etal_acl_2015} extend their previous work by introducing a new transition action to get better performance. ~\citet{damonte_etal_eacl_2017} propose a complete transition-based approach that parses sentences left-to-right in linear time. The recent neural AMR parsing could be roughly grouped into two categories. On the one hand, the generic seq2seq-based approaches have been widely used for AMR parsing which show competitive performance~\cite{peng_etal_eacl_2017,noord_bos_2017,konstas_etal_acl_2017,ge_etal_ijcai_2019}. On the other hand, to better model the graph structure on the target side, graph-based models are well studies for AMR parsing which achieve the state-of-the-art-performance~\cite{lyu_titov_acl_2018,guo_lu_emnlp_2018,groschwitz_etal_acl_2018,zhang_etal_acl_2019,zhang_etal_emnlp_2019,cai_lam_acl_2020}.

\section{Conclusion}
\label{sect:conclusion}
In this paper we proposed a seq2seq-based pre-training approach to improving the performance of seq2seq-based AMR parsing. To this end, we designed three relevant seq2seq learning tasks, including machine translation, syntactic parsing, and AMR parsing itself.  Then we built seq2seq pre-trained models through either single or joint pre-training tasks. Detail experimentation shows that both the single and joint pre-trained models substantially improve our baseline and the performance reaches the state of the art. The accomplishment is encouraging since we achieve this simply by using the generic seq2seq framework rather than complex models.

\section*{Acknowledgments}
We thank the anonymous reviewers for their insightful comments and suggestions. This work is supported by the National Natural Science Foundation of China (Grant No. 61876120)\ignore{, and the Priority Academic Program Development of Jiangsu Higher Education Institutions}.

\bibliography{emnlp2020}
\bibliographystyle{acl_natbib}
\end{document}